\documentclass[lettersize,journal]{IEEEtran}
\usepackage{amsmath,amsfonts}
\usepackage{algorithmic}
\usepackage{algorithm}
\usepackage{array}
\usepackage[caption=false,font=normalsize,labelfont=sf,textfont=sf]{subfig}
\usepackage{textcomp}
\usepackage{stfloats}
\usepackage{xspace}
\usepackage{url}
\usepackage{verbatim}
\usepackage{graphicx}
\usepackage{booktabs}
\usepackage[colorlinks,linkcolor=blue]{hyperref}
\usepackage{cleveref}
\usepackage{cite}
\usepackage{orcidlink}
\usepackage{threeparttable}
\usepackage{multirow}
\usepackage{tabularx}
\newcolumntype{C}{>{\centering\arraybackslash}X} 
\newcommand{\ie}{{\emph{i.e.}},\xspace}

\newcommand{\etal}{{\emph{et al.}}}

\begin{document}

\title{ScaleDepth: Decomposing Metric Depth Estimation into Scale Prediction and Relative Depth Estimation} 

\author{Ruijie Zhu~\orcidlink{0000-0001-6092-0712}, 
    Chuxin Wang~\orcidlink{0000-0003-1431-7677},
    Ziyang Song~\orcidlink{0009-0009-6348-8713}, 
    Li Liu~\orcidlink{0009-0004-3280-8490},\\
    Tianzhu Zhang~\orcidlink{0000-0003-0764-6106},~\IEEEmembership{Member,~IEEE,}
    Yongdong Zhang~\orcidlink{0000-0002-1151-1792},~\IEEEmembership{Senior Member,~IEEE}
\thanks{The authors are with School of Information Science and Technology, University of Science and Technology of China (USTC), Hefei 230026, China. 
(e-mail: 
\{ruijiezhu, wcx0602, songziyang, liu\_li\}@mail.ustc.edu.cn;
\{tzzhang, zhyd73\}@ustc.edu.cn). Corresponding author: Tianzhu Zhang.}
\thanks{This work has been submitted to the IEEE for possible publication. Copyright may be transferred without notice, after which this version may no longer be accessible.}
}



\maketitle

\begin{abstract}
Estimating depth from a single image is a challenging visual task. 
Compared to relative depth estimation, metric depth estimation attracts more attention due to its practical physical significance and critical applications in real-life scenarios.
However, existing metric depth estimation methods are typically trained on specific datasets with similar scenes, facing challenges in generalizing across scenes with significant scale variations.
To address this challenge, we propose a novel monocular depth estimation method called ScaleDepth. 
Our method decomposes metric depth into scene scale and relative depth, and predicts them through a semantic-aware scale prediction (SASP) module and an adaptive relative depth estimation (ARDE) module, respectively.
The proposed ScaleDepth enjoys several merits.
First, the SASP module can implicitly combine structural and semantic features of the images to predict precise scene scales.
Second, the ARDE module can adaptively estimate the relative depth distribution of each image within a normalized depth space.
Third, our method achieves metric depth estimation for both indoor and outdoor scenes in a unified framework, without the need for setting the depth range or fine-tuning model.
Extensive experiments demonstrate that our method attains state-of-the-art performance across indoor, outdoor, unconstrained, and unseen scenes.
Project page: \url{https://ruijiezhu94.github.io/ScaleDepth}.
\end{abstract}


\section{Introduction}
\label{sec:intro}
\IEEEPARstart{D}{epth} estimation is a fundamental task in 3D vision, which has vital applications in many downstream tasks, such as autonomous driving~\cite{schon2021mgnet}, augmented reality~\cite{yucel2021real}, and 3D reconstruction~\cite{yang2020mobile3drecon}.
Compared with multi-view stereo, Single-Image Depth Estimation (SIDE) has gained widespread attention for its cost-effectiveness and ease of deployment.
%
Due to the lack of geometric constraints across multiviews, SIDE is an ill-posed problem.
As a result, traditional methods~\cite{saxena2005learning,michels2005high, hoiem2007recovering, saxena2008make3d} heavily rely on hand-crafted features to learn geometric priors.
%

To address this issue, a series of learning-based methods~\cite{eigen2014depth, eigen2015predicting, liu2015deep, wang2015towards, zhou2015learning, fu2018deep, laina2016deeper, chen2016single} have been proposed, which can be broadly categorized into Relative Depth Estimation (RDE) and Metric Depth Estimation (MDE) methods. 
RDE methods~\cite{zoran2015learning, xian2020structure, Ranftl2022, yang2024depth, ke2024repurposing} aim to infer the relative depth relationship between objects in a scene, which is independent of the scale. 
However, these methods may struggle to address complex real-world applications, such as robot grasping and obstacle avoidance. 
Therefore, MDE methods~\cite{piccinelli2023idisc, Ning_2023_ait, zhao2023unleashing, shao2023nddepth} have become the mainstream in monocular depth estimation methods.
Typically, they are trained on a single dataset with image-depth pairs and directly regress per-pixel metric depth maps.
However, these methods ignore scenes with significant scale differences and are difficult to directly generalize from indoors to outdoors.
Some methods~\cite{bhat2021adabins, li2022binsformer, bhat2022localbins, shao2023IEBins, zhu2023habins} discretize depth into bins and adaptively estimate the depth distribution for each image, but they also neglect the scale differences between indoor and outdoor MDE.
Recently, Zoedepth~\cite{bhat2023zoedepth} attempts to overcome this problem, but still requires separate prediction heads to handle indoor and outdoor depth predictions individually.
Some other methods~\cite{guizilini2023towards, yin2023metric3d} use camera parameters to resolve scale amibiguity and fuse numerous datasets from various scenarios for model training, but they also lack explicit modeling of scene scale and rely on large amounts of training data.

\begin{figure*}[ht]
    \centering  
    \subfloat[Outdoor scene]{\label{fig:outdoor}\includegraphics[width=0.2\linewidth]{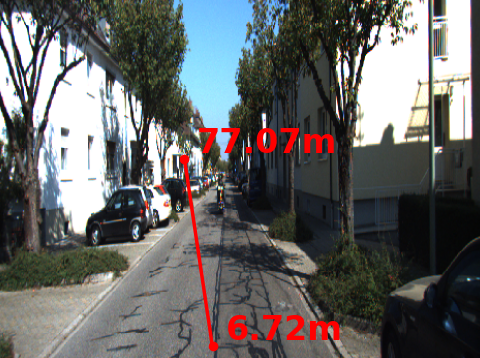}}
    \subfloat[kitchen]{\label{fig:kitchen}\includegraphics[width=0.2\linewidth]{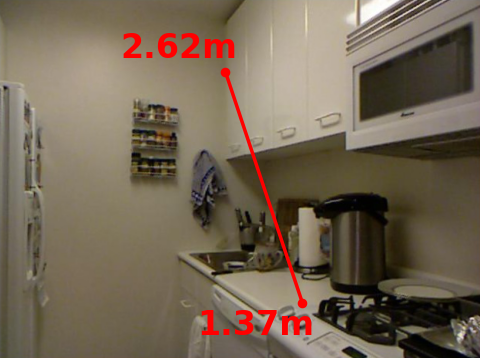}}
    \subfloat[Classroom1]{\label{fig:classroom1}\includegraphics[width=0.2\linewidth]{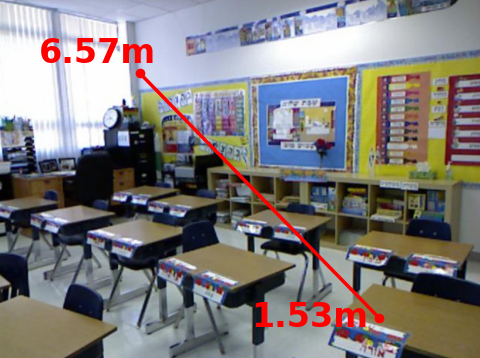}}
    \subfloat[Classroom2]{\label{fig:classroom2}\includegraphics[width=0.2\linewidth]{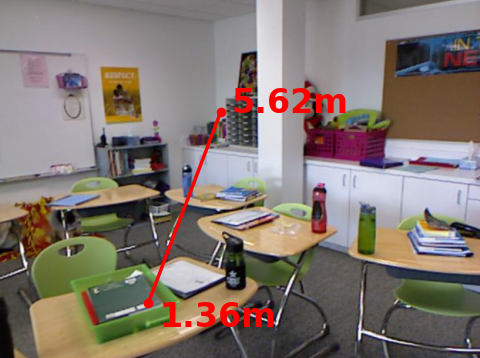}}
    \subfloat[Plants]{\label{fig:plant}\includegraphics[width=0.2\linewidth]{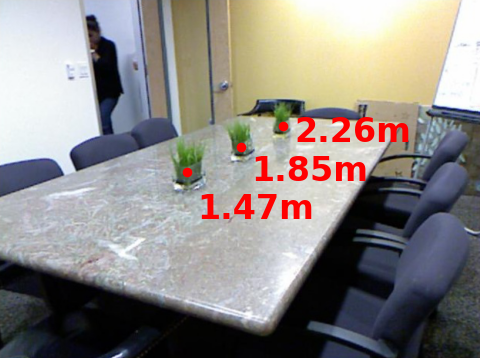}}
    \caption{\textbf{Examples of various scenes and objects with different depths.
    } Scenes of different categories typically exhibit large scale variations (a, b, and c), while scenes of the same category have similar scales (c and d). Same objects  have varying depths within the same scene due to their different placement (e).}
    \label{Fig.main}
\end{figure*}

Based on the above discussion, we observe that existing methods usually overlook the impact of scene scale on MDE, leading to challenges in predicting depth accurately in scenes with diverse depth ranges.
To address this issue, we summarize two key points that require further consideration.
(1) \textbf{Scenes from different categories usually exhibit larger differences in depth range, while scenes from the same category may have smaller differences.}
For example, the depth ranges of the outdoor scene, the kitchen and the classroom exhibit significant disparities (\Cref{fig:outdoor,,fig:kitchen,,fig:classroom1}).
And the depth ranges corresponding to different classrooms are similar (\Cref{fig:classroom1,,fig:classroom2}).
Therefore, we argue that scale difference is the primary reason that prevents existing MDE methods from unifying indoor and outdoor depth estimation. 
If we can explicitly model the scale of the scene, the model only needs to focus on inferring relative depth relationships. 
However, relying solely on scene category information to infer the scale is insufficient, as it is additionally influenced by the own structure of the scene.
Therefore, it is crucial to consider both the structural and semantic information of each scene for scale prediction.
(2) \textbf{In a certain scene, even objects of the same category may have different depths.}
As shown in \Cref{fig:plant}, the plants on the table have different depths depending on their placement.
If there is only one pot of plant on the table, the model may struggle to accurately estimate depth solely by extracting its features.
In that case, it is necessary to leverage the surrounding objects, \ie the tabletop, to assist depth prediction.
Therefore, we argue that aggregating features from depth-related regions can benefit local structure modeling and inferences of relative depth relationships.

\begin{figure*}[ht] 
  \centering 
  \includegraphics[width=1\textwidth]{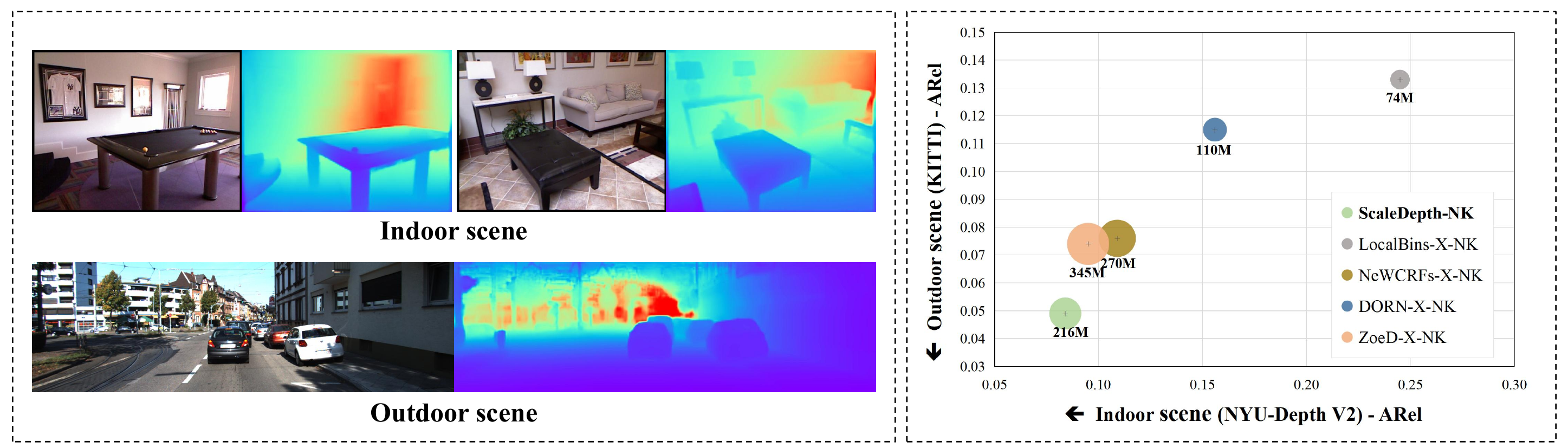} 
  \caption{\textbf{Within a unified framework, our method ScaleDepth achieves both accurate indoor and outdoor metric depth estimation without setting depth ranges or finetuning models.} Left: the input RGB image and corresponding depth prediction. Right: the comparison of model parameters and performance. With overall fewer parameters, our model ScaleDepth-NK significantly outperforms the state-of-the-art methods under same experimental settings.
  } 
  \label{fig:moti2} 
\end{figure*}

Based on the above analysis, we propose a novel depth estimation method by decomposing MDE into scale prediction and relative depth estimation, achieving both accurate indoor and outdoor metric depth estimation in a unified framework (see~\Cref{fig:moti2}).
Our method mainly consists of a Semantic-Aware Scale Prediction (SASP) module and an Adaptive Relative Depth Estimation (ARDE) module. 
\textbf{In the SASP module}, we design scale queries for scale prediction and leverage text-image feature similarity to impose semantic constraints.
To incorporate structural information from the scene, we aggregate image features through scale queries.
To integrate semantic information from the scene, we formulate text prompts and encode scene categories into text embeddings by the frozen CLIP~\cite{radford2021learning} text encoder.
By calculating the similarity between scale queries and text embeddings, we construct effective semantic constraints with the scene category labels.
%
%
\textbf{In the ADRE module}, we adopt a set of bin queries to aggregate features in depth-related regions and predict relative depth maps in a discrete regression manner.
Specifically, we utilize bin queries to interact with image features through mask attention and predict bins within a normalized depth space.
The probability of pixels belonging to bins is then predicted to weight bin centers for the relative depth estimation.
Finally, the relative depth map is multiplied by the predicted scale to obtain the metric depth map.
%

The main contributions of this paper can be summarized as follows.
(1) We propose ScaleDepth, which can infer accurate metric depth across scenes with significant scale differences within a unified framework.
(2) The proposed SASP module implicitly integrates semantic and structural information to predict the scale of each scene, while the ARDE module adaptively indicates the relative depth distribution for each image.
(3) Extensive experiments demonstrate that our model reaches the state of the art on indoor and outdoor benchmarks and exhibits satisfactory generalization in zero-shot evaluations.

\section{Related Work}
\label{sec:relatedwork}

In this section, we provide a brief overview of related work in metric depth estimation and relative depth estimation.

\textbf{Metric Depth Estimation (MDE). }
Continuous regression-based MDE is the most classical paradigm in MDE.
Eigen~\etal~\cite{eigen2014depth} first introduce convolution neural networks for end-to-end training in MDE.
Subsequently, numerous works have been proposed, primarily concentrating on the following directions:
(a) Improving network architecture, including approaches like residual networks~\cite{laina2016deeper, hu2018squeeze}, multi-scale fusion~\cite{eigen2015predicting, wang2019web, zhang2018hard, xu2021multi}, transformers~\cite{ranftl2021vision, yuan2022new} and diffusion models~\cite{zhao2023unleashing, ji2023ddp}.
(b) Designing novel loss functions or constraints, such as planar, normal, and gradient constraints~\cite{lee2019big, li2021structdepth, yu2020p, li2017two,yin2019enforcing}.
(c) Leveraging auxiliary information or multi-task learning, for example, surface normal estimation~\cite{qi2018geonet} and semantic segmentation~\cite{xu2018pad, Ning_2023_ait}.
Recently, the focus of MDE methods has shifted towards discrete regression.
In contrast to continuous regression, discrete regression-based methods discretize depth into classes and predict the corresponding classes of pixels.
Initially, Cao~\etal~\cite{cao2017estimating} transform depth estimation into a classification problem.
Afterwards, Fu~\etal~\cite{fu2018deep} propose ordinal regression, suggesting depth discretization with increasing space and designing loss to consider the ordinal correlation of depth values.
Subsequent methods~\cite{bhat2021adabins, bhat2022localbins, li2022binsformer, shao2023IEBins, bhat2023zoedepth} adopt a classification-regression mixed paradigm and design adaptive bin division strategies.
%
Among them, Zoedepth~\cite{bhat2023zoedepth} proposes an automatic routing strategy for joint indoor and outdoor depth estimation, which adaptively selects the corresponding MDE prediction head to estimate depth.
Besides, there are also some MDE methods~\cite{guizilini2023towards, yin2023metric3d} focus on joint training on multiple datasets.
However, they mainly focus on solving the scale ambiguity caused by different camera parameters without explicitly modeling the scene scale.
Unlike the above methods, the proposed ScaleDepth decomposes MDE into scale prediction and RDE, enabling depth prediction for scenes with different depth ranges within a unified framework.

\textbf{Relative Depth Estimation (RDE).}
Different with MDE, RDE focuses on pairwise depth order prediction~\cite{zoran2015learning} or the depth ordering of all pixels~\cite{xian2020structure}.
The advantage of RDE methods is evident: the ordinal relationships of pixels are independent of the scene scale, making the model more accessible to generalize to scenes with significant scale differences.
Zoran~\etal~\cite{zoran2015learning} and Zhou~\etal~\cite{zhou2015learning} first introduce a deep neural network to estimate pairwise depth order and implement intrinsic image decomposition.
Subsequent methods focus mainly on boosting the MDE performance of the model with the help of RDE.
Chen~\etal~\cite{chen2016single} study the use of relative depth annotation to predict the metric depth of a single image taken in the unconstrained settings.
Eigen~\etal~\cite{eigen2014depth} and its follow-ups~\cite{li2018megadepth, Ranftl2022} use scale-invariant loss and its variants to alleviate scale ambiguity in depth estimation.
Jun~\etal~\cite{jun2022depth} propose a metric depth decomposition with separate decoders, which predicts relative depth maps and metric depth maps, respectively, to reduce the dependence on the metric depth annotation.
Recently, some methods have attracted wide attention from community due to their impressive generalization ability, such as Marigold~\cite{ke2024repurposing} and Depth Anything~\cite{yang2024depth}.
However, they only predict relative depth maps, which still requires extra scale and shift factors to convert to metric depth.
Compared with them, our method can adaptively estimate metric depth maps across multiple datasets and maintain satisfactory generalization ability even in scenes with large scale variations.

\section{Methodology}
\label{sec:method}

\begin{figure*}[t]
    \centering
    \includegraphics[width=\linewidth]{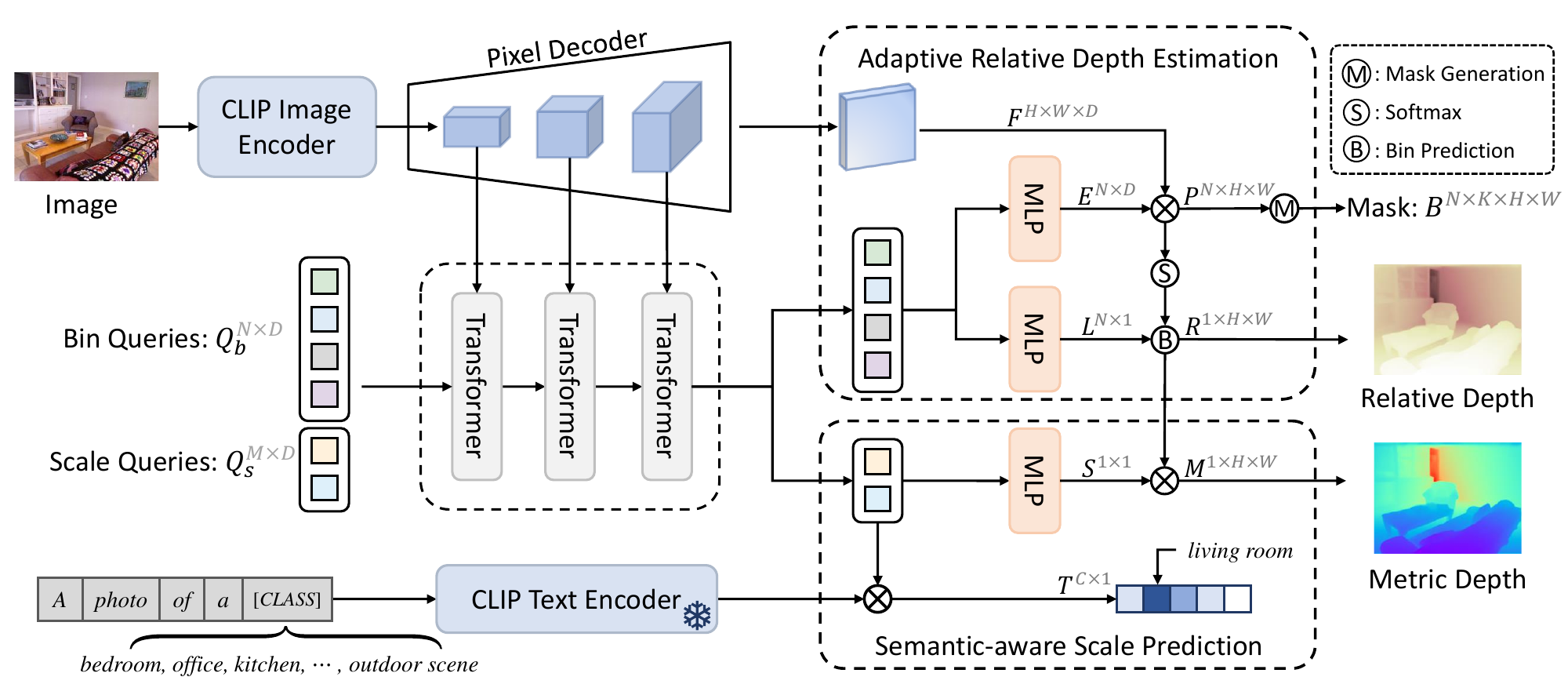}
    \caption{
        \textbf{The overall architecture of the proposed ScaleDepth.} 
        We design bin queries to predict relative depth distribution and scale queries to predict scene scale. During training, we preset text prompts containing 28 scene categories as input to the frozen CLIP text encoder. We then calculate the similarity between the updated scale queries and text embedding, and utilize the scene category as its auxiliary supervision. During inference, only a single image is required to obtain the relative depth and scene scale, thereby synthesizing a metric depth map.
    }
    \label{fig:framework}
\end{figure*}

In this section, we present our method by decomposing the metric depth estimation into Semantic-Aware Scale Prediction and Adaptive Relative Depth Estimation.
The overall architecture is illustrated in~\Cref{fig:framework}.

\subsection{Overview}
As shown in~\Cref{fig:framework}, our method mainly consists of a semantic-aware scale prediction (SASP) module and an adaptive relative depth estimation (ARDE) module.
Given an input image $I$, we first extract multilevel image features $\{F_l\}_{l=0}^3$ via a CLIP image encoder~\cite{pmlr-openclip}.
In the pixel decoder, we project and flatten the image features into $D$-dimensional vectors $F \in \mathbb{R}^{H W \times D}$ at each level, which are then sent to the transformer layers for interaction with the queries.
To decompose MDE into scale prediction and RDE, we design two sets of object queries. 
The first one is referred as scale queries $Q_s \in \mathbb{R}^{M\times D}$, corresponding to the scale prediction,
while the other is termed bin queries  $Q_b \in \mathbb{R}^{N\times D}$, corresponding to the relative depth estimation.
In the transformer layers, we adopt masked attention inspired by Mask2Former~\cite{cheng2021mask2former} to obtain the updated queries $\widehat{Q}_s$ and $\widehat{Q}_b$.
Then the queries $\widehat{Q}_s$ and $\widehat{Q}_b$ are passed through the SASP module and the ARDE module to generate a scale factor $S \in \mathbb{R}^{1 \times 1}$ and a relative depth map $R \in \mathbb{R}^{1 \times H \times W}$, respectively.
Finally, the metric depth predictions $M \in \mathbb{R}^{1 \times H \times W}$ are obtained by directly multiplying scale $S$ and relative depth map $R$.
During training, we employ a frozen CLIP text encoder branch as auxiliary supervision of the SASP module, while this branch is not required during inference.

\subsection{Adaptive Relative Depth Estimation Module}
To adaptively predict the relative depth map for each image, we adopt a bin-based mechanism in the proposed ARDE module.
In contrast to previous methods~\cite{bhat2021adabins, li2022binsformer}, we define a normalized depth space and partition it into bins within the 0-1 depth range. 
When each bin represents a depth class, we calculate the similarity between image features and bin features to classify all pixels on these bins.
After predicting the classification probabilities, the relative depth map is calculated by weighting the central depth of the bins.
At each transformer layer, we use the updated bin queries to generate a relative depth prediction.
Through mask generation, the binary attention masks are generated simultaneously.
We send these masks into the next transformer layer to enable the attention interaction between bin queries and image features in depth-related regions.
The proposed ADRE module predicts scale-invariant depth and allows the model to aggregate image features adaptively across different depth ranges.

\textbf{Bin Prediction.}
We first adaptively estimate a group of bins to discretize depth into classes for each image.
Given the output bin queries $\widehat{Q}_b \in \mathbb{R}^{N\times D}$, we pass them through independent MLPs to predict bin lengths $L \in \mathbb{R}^{N \times 1}$ and bin features $E \in \mathbb{R}^{N \times D}$.
Then we normalize bin lengths and calculate the central value of the $i$-th bin as
\begin{equation}
    \theta_i =\frac{1}{2} L_i +\sum _{j=1}^{i-1} L_j,
\end{equation}
where $L_i$ denotes the length of the $i$-th bin.
We use the central values of the bins to represent the classes of the relative depth within the 0-1 depth space.

\textbf{Mask Generation.}
To adaptively aggregate the features of depth-related regions, we generate attention masks for the next-layer attention.
The feature similarity between bin features and image features $P \in \mathbb{R}^{N \times H \times W}$ can then be calculated by matrix multiplication:
\begin{equation}
    P =  {E} \times F^{\top},
\end{equation}
where $E \in \mathbb{R}^{N\times D}$ denotes the bin features and $F \in \mathbb{R}^{H W \times D}$ denotes the image features extracted by the pixel decoder.
Then, the attention masks of the next layer $B \in \mathbb{R}^{N \times K \times H \times W}$ are generated by resizing (to match the size of image features at the next layer), repeating ($K$ times to match the number of attention heads), and binarizing (thresholded at 0.5) the feature similarity $P \in \mathbb{R}^{N \times H \times W}$.
The updated attention masks help the model focus on the features activated by the discrete bins, facilitating the aggregation of depth-related image features for adaptive learning and fast convergence.
Please note that we only generate attention masks for bin
queries and not for scale queries, since each bin query aims to aggregate image features in a depth-related region and scale queries aim to perceive global structural information of the scene.

\textbf{Relative Depth Estimation.}
To obtain the relative depth map $R$, we calculate the per-pixel classification probabilities on bins and use them to weight bin centers as 
\begin{equation}
    R = {\theta}^{\top} \times \mathrm{softmax}(P),
\end{equation}
where $\theta \in \mathbb{R}^{ N \times 1}$ is the bin centers and the operation $\mathrm{softmax}$ calculates the probabilities of pixels falling into $N$ bins.
The predicted relative depth map $R$ is a normalized depth map in 0-1 depth space, independent of scale.
This design allows the model to focus on estimating the relative depth distribution between pixels, mitigating the impact of scale differences on metric depth estimation.

\subsection{Semantic-Aware Scale Prediction Module}
To achieve semantic-aware scale prediction, we leverage the scene category to guide the model in capturing global semantic information. 
In the previous work VPD~\cite{zhao2023unleashing}, the scene category is directly used as input to the model, preventing it from generalizing to scenes of unknown categories. 
In contrast, the proposed SASP module utilizes image-text similarity as a constraint, aligning the scale queries with the text embedding of its corresponding scene category. 
This approach encourages scale queries to learn global semantic features from the image, enabling the model to generalize to scenes of unknown categories without relying on scene category information. 
Additionally, compared with directly using text embedding to predict the scale, the scale queries aggregate the features extracted by the CLIP image encoder for scale prediction.
Benefiting from the inherent alignment of image and text features in the CLIP model, the proposed SASP module implicitly combines the structural and semantic information of the scenes, allowing the model to predict the scene scale for each image adaptively.

\textbf{Text-image Similarity Calculation.} 
To guide the model in capturing semantic information, we first generate text prompts through manually crafted text templates.
Given $C$ class names, the text prompts $\{t_i\}_{i=1}^{C}$ is in forms like ``a photo of a [CLASS]".
Then, the text embeddings $F_t \in \mathbb{R}^{C \times D_t}$ of $C$ scene categories are derived by passing the text prompts into the frozen CLIP text encoder.
Given the output scale queries $\widehat{Q}_s \in \mathbb{R}^{M\times D}$, we concatenate and project them to a feature vector ${F}_c \in \mathbb{R}^{1\times D_t}$.
The text-image similarity $T$ of class $i$ is then calculated as
\begin{equation}
    T_i=\frac{ \exp(\cos \langle F_t^i,F_c \rangle /\tau )}{ {\sum_{j=1}^{C}}\exp(\cos \langle F_t^j,F_c \rangle /\tau)},
\end{equation}
when $\cos \langle \cdot \rangle$ calculates the cosine similarity between features, $F_{t}^{i}$ denotes the text embedding of class $i$, and $\tau$ is the temperature parameter.
The computed image-text similarity represents the classification probabilities of the image in different scenes and can be supervised by scene category information during training.
In our experiments, we present the number of scene categories $C$ to 28.

\textbf{Scale Prediction and Metric Depth Estimation.}
After obtaining the semantic-aware scale queries, we directly project them to the scale factor $S \in \mathbb{R}^{1 \times 1}$ by an MLP. 
Finally, the metric depth map $M \in \mathbb{R}^{1 \times H \times W}$ is calculated by multiplying scale $S$ and relative depth map $R$.
By explicitly modeling the scale of the scene, our method combines the advantages of direct and discrete regression-based methods, enabling depth prediction within an unfixed depth range while adapting to the depth distribution of each image.
This paradigm allows our method to estimate accurate metric depth maps in both indoor and outdoor scenes without pretraining on extensive depth datasets.

\subsection{Loss Function}
To effectively constrain the predicted relative depth map and scale, we reimplement the Scale-Invariant (SI) loss introduced by Eigen~\etal~\cite{eigen2014depth}. 
Besides, we impose an additional Text-Image (TI) similarity loss to supervise semantic-aware scale prediction for the scale queries.

\textbf{Pixel-wise Depth Loss.}
We rewrite SI loss as two terms:
\begin{equation}
    \mathcal{L}_{\text{SI}} =\alpha \sqrt{\mathbb{V}[\delta] + \lambda \mathbb{E}^2[\epsilon]},
\end{equation}
where $\delta =\log\overline{M} -\log R$, $\epsilon =\log\overline{M} -\log M$, $\overline{M}$ denotes the ground truth depth.
When $\mathbb{V}[\delta]$ and $\mathbb{E}[\epsilon]$ respectively compute the variance and expectation of the log depth errors for all valid pixels. 
The former is scale-invariant, while the latter takes scale into account.
In our experiments, we set $\alpha$ to 10 and $\lambda$ to 0.15 as customary.

\textbf{Text-image Similarity Loss.}
We use the cross-entropy loss to provide semantic supervision of the scale queries:
\begin{equation}
    \mathcal{L}_{\text{TI}}  = -\sum_{i=1}^{C} c_i\log(T_i),
\end{equation}
where $\{{c_i}\}_{i=1}^C$ is the one-hot scene category label and $T_i$ is the text-image similarity of the $i$-th scene category.
%

Finally, we define the total loss as:
\begin{equation}
    \mathcal{L}_{\text{total}} =  \mathcal{L}_{\text{SI}} + \beta \mathcal{L}_{\text{TI}},
\end{equation}
where $\beta$ is set as 0.01 in our experiments.

\section{Experiments}
\label{sec:Experiments}

\subsection{Datasets}
\noindent\textbf{NYU-Depth V2} is an indoor dataset with RGB images and corresponding depth maps at a $480\times640$ resolution. 
Following the official split, we use 24231 image-depth pairs for training and 654 images for testing.

\noindent\textbf{KITTI} is an outdoor dataset collected by the equipment mounted on a moving vehicle. 
Following KBCrop~\cite{Uhrig2017THREEDV}, all the RGB images and depth maps are cropped to a resolution of $1216\times352$. 
We adopt the Eigen split~\cite{eigen2014depth} with 23158 training images and 652 test images to train and evaluate our method. Besides, the capturing depth range of the Eigen split is 0-80m.

\noindent\textbf{Eight Unseen Datasets} are introduced in our experiments for zero-shot evaluation.
We use SUN RGB-D~\cite{song2015sunrgbd}, iBims-1 Benchmark~\cite{koch2018ibims}, DIODE Indoor~\cite{vasiljevic2019diode} and HyperSim~\cite{roberts2021hypersim} for indoor evaluation, and use Virtual KITTI 2~\cite{cabon2020virtual}, DDAD~\cite{guizilini20203d}, DIML Outdoor~\cite{cho2021diml} and DIODE Outdoor~\cite{vasiljevic2019diode} for outdoor evaluation.
No depth range of these datasets is given on zero-shot evaluations of our models.

\subsection{Implementation Details}

\noindent\textbf{The Proposed Models.}
For distinction, we denote the proposed models as ScaleDepth-\{dataset\}.
The dataset refers to the datasets used for training, which includes ``N'' (NYU-Depth V2), ``K'' (KITTI), and ``NK'' (both NYU-Depth V2 and KITTI).
All the proposed models are trained without setting depth ranges and tested without any finetuning. 

\noindent \textbf{Training details. }
The proposed ScaleDepth is implemented in PyTorch. 
We use the AdamW optimizer~\cite{loshchilov2017decoupled} $(\beta_1, \beta_2, wd)=(0.9, 0.999, 0.05)$ with an initial learning rate of 1e-4. 
All the models are trained for 40000 iterations on 4 NVIDIA RTX 3090 GPUs. The batch size of ScaleDepth-N and ScaleDepth-K is set to 24, and the batch size of ScaleDepth-NK is set to 32. 
The total training of the model takes approximately 8-10 hours.
During training on a single dataset, we randomly crop images to $480\times480$ for NYU-Depth V2 and to $352\times1120$ for KITTI. During joint training on both datasets, we randomly crop images to $352\times512$.

\noindent \textbf{Evaluation details.}
We use the standard five error metrics and three accuracy metrics for evaluation.
Specifically, the error metrics include absolute mean relative error (ARel), log error (${\rm log}_{10}$), scale-invariant log error (SILog), root mean squared error (RMSE) and its log variant (RMSL).
The accuracy metrics include the percentage of inlier pixels $\delta$ for three thresholds, i.e. ($ \delta_1 < 1.25 $, $ \delta_2 < 1.25^2 $, $ \delta_3 < 1.25^3 $).
During evaluations, we set the maximum valid depth for NYU-Depth V2 to 10m, and for KITTI to 80m.
Notably, we only set the maximum depth to identify valid pixels, which is unnecessary during training.
\begin{figure*}[htbp]
  \centering
   \includegraphics[width=0.95\linewidth]{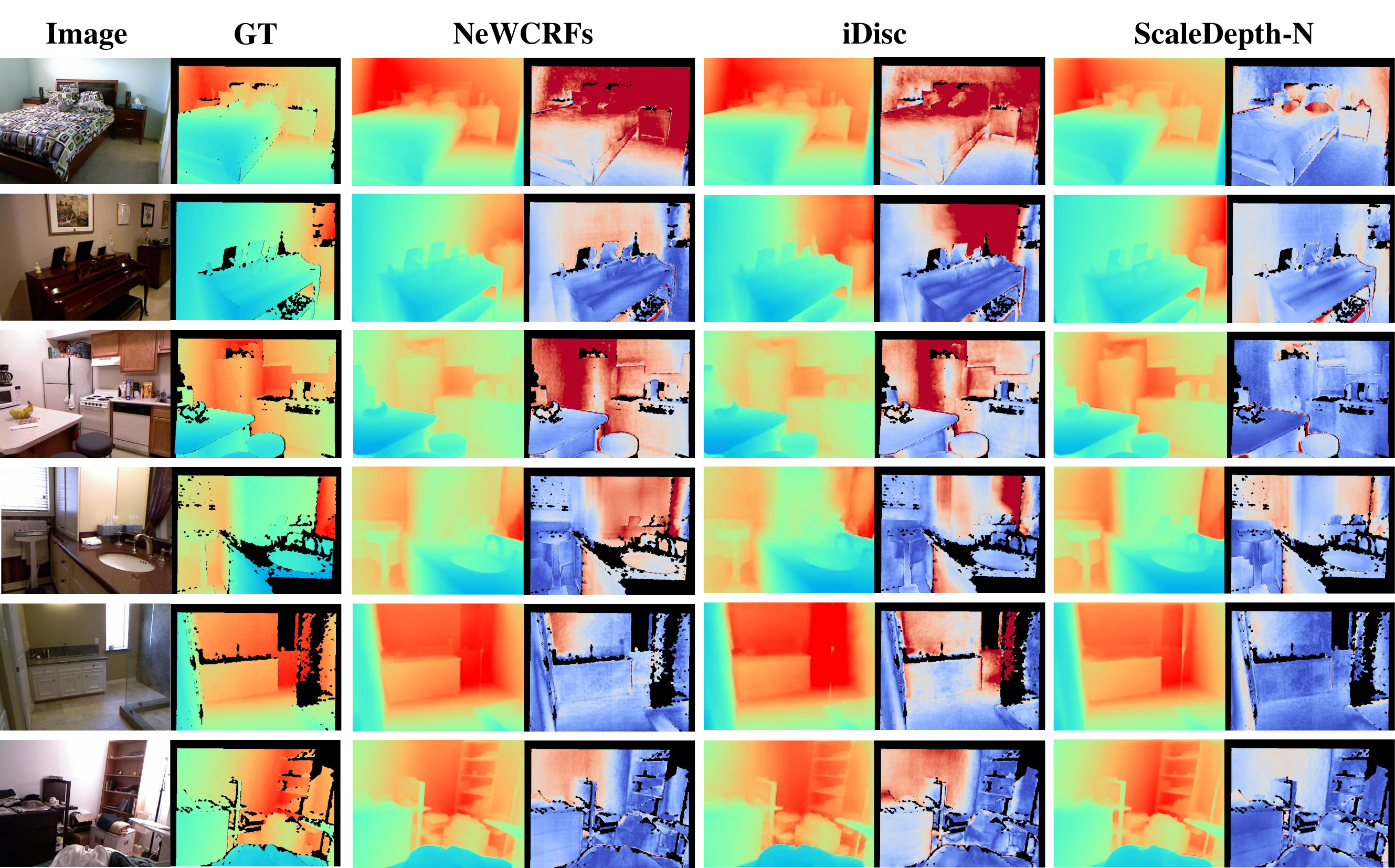}
   \caption{\textbf{The qualitative comparison on NYU-Depth V2 dataset.} For each test sample pair, the left is the depth map and the right is the error map. In each map, blue corresponds to lower \textbf{(metric depth or error) values} and red to higher values.}
   \label{fig:nyu}
    \vspace{1em}
    \centering
    \includegraphics[width=0.96\linewidth]{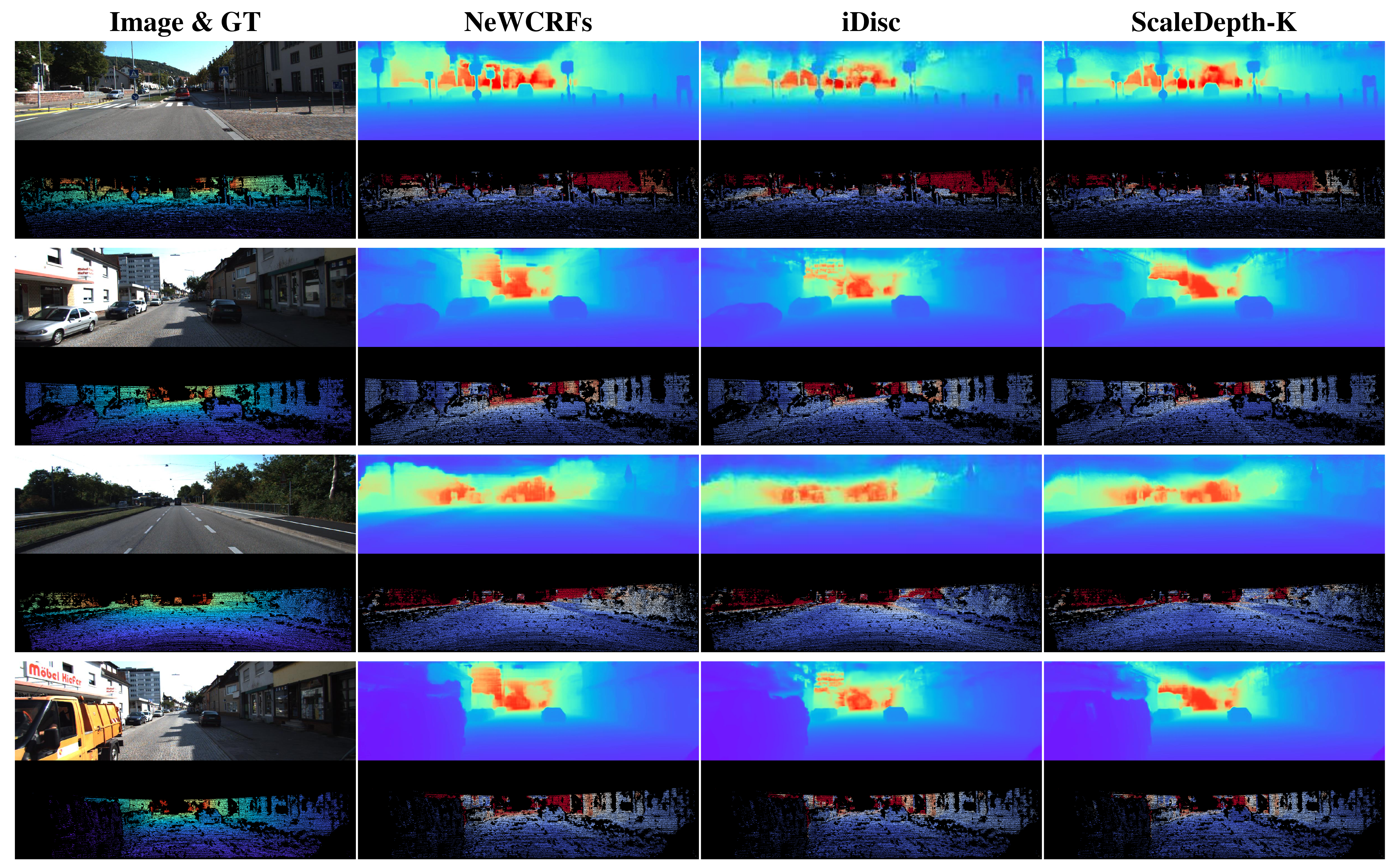}
    \caption{\textbf{The qualitative comparison on KITTI dataset. } Each block with 8 images corresponds to a test sample. The first column displays the original image and its corresponding depth ground truth. The subsequent columns show the predicted depth maps and error maps of the corresponding models. We use color map \textit{rainbow} and \textit{coolwarm} to map the depth and error values respectively, and the error values are mapped using a unified range of 0-5m. Blue corresponds to lower \textbf{(metric depth or error) values} and red to higher values.
    }
   \label{fig:kitti_compare}
\end{figure*}

\subsection{Comparison with the State of the Art}
In this section, we compare our method with state-of-the-art approaches in indoor, outdoor, unconstrained, and unseen scenes, respectively. 

\begin{table}[t]
\centering
\caption{
\textbf{Quantitative results on NYU-Depth V2 dataset.} 
The best results are in \textbf{bold}, and the second best are \underline{underlined}. 
}
\label{table:nyu}
\resizebox{\linewidth}{!}{
\begin{tabularx}{1.1\linewidth}{@{}l|c|*{6}{C}@{}}
        \toprule
            Method & Params & ARel$\downarrow$ & RMSE$\downarrow$ & ${\rm log}_{10}\downarrow$ &   $\delta_1\uparrow$ & $\delta_2 \uparrow$ & $\delta_3 \uparrow$ \\
        \midrule
        
        DORN~\cite{fu2018deep}  & 110M & 0.115 &  0.509 &  0.051 & 0.828 & 0.965 & 0.992 \\
        BTS~\cite{lee2019big}  & 47M & 0.110 & 0.392 & 0.047 & 0.885 & 0.978 & 0.994 \\
        AdaBins~\cite{bhat2021adabins}  & 78M & 0.103 &  0.364 & 0.044 & 0.903 & 0.984 & 0.997 \\
        NeWCRFs~\cite{yuan2022new}  & 270M & 0.095 & 0.334 &  0.041 & 0.922 & 0.992 & 0.998 \\
        BinsFormer~\cite{li2022binsformer} & 255M & 0.094 & 0.330 & 0.040 & 0.925 & 0.989 & 0.997 \\
        iDisc~\cite{piccinelli2023idisc}  & 209M & {0.086} & {0.313} & {0.037} & {0.940} & {0.993} & \textbf{0.999} \\
        IEBins~\cite{shao2023IEBins} & 273M & 0.087 & 0.314 & 0.038 & 0.936 & 0.992 & 0.998 \\
        VPD~\cite{zhao2023unleashing} & {872M} & \textbf{0.069} & \textbf{0.254} & \textbf{0.030} & \textbf{0.964} & \textbf{0.995} & \textbf{0.999} \\
        \textbf{ScaleDepth-N} & 216M & \underline{0.074} & \underline{0.267} & \underline{0.032} & \underline{0.957} & \underline{0.994} & \textbf{0.999} \\
        \bottomrule
    \end{tabularx}
}
\end{table}

\begin{table}[t]
\centering
\caption{
\textbf{Quantitative results on the Eigen split of KITTI dataset.}
Measurements are made for the depth range from 0-80m. 
}\label{table:kitti} 
\begin{threeparttable}
\resizebox{\linewidth}{!}{
    \begin{tabularx}{1.1\linewidth}{@{}l|*{7}{C}@{}}
        \toprule
        Method & ARel$\downarrow$ & RMSE$\downarrow$ & RMSL$\downarrow$ & SRel$\downarrow$ & $\delta_1\uparrow$ & $\delta_2 \uparrow$ & $\delta_3 \uparrow$ \\
        \midrule
        DORN~\cite{fu2018deep} & 0.072 & 2.727 & 0.120  & 0.307 & 0.932 & 0.984 & 0.994   \\
        BTS~\cite{lee2019big}  & 0.060 & 2.798  & 0.096 & 0.249 & 0.955 & 0.993 & 0.998 \\
        AdaBins~\cite{bhat2021adabins}  & 0.060 & 2.372 & 0.090 & 0.197 & 0.963 & 0.995 & \underline{0.999}  \\    
        NeWCRFs~\cite{yuan2022new}  & 0.052 & 2.129  & 0.079  & 0.155 & 0.974 & 0.997 & \underline{0.999} \\
        BinsFormer~\cite{li2022binsformer}  & 0.052 & 2.098  & 0.079 & 0.151 & 0.974 & 0.997 & \underline{0.999}\\
        iDisc~\cite{piccinelli2023idisc} & \underline{0.050} & 2.067 & 0.077 & 0.145 & 0.977 & 0.997 & \underline{0.999} \\
        IEBins~\cite{shao2023IEBins} & \underline{0.050} & \underline{2.011} &  \underline{0.075}& \underline{0.142} & \underline{0.978} & \textbf{0.998} & \underline{0.999} \\
        \textbf{ScaleDepth-K} & \textbf{0.048} & \textbf{1.987} & \textbf{0.073} & \textbf{0.136} & \textbf{0.980} & \textbf{0.998} & \textbf{1.000}  \\
        \bottomrule
    \end{tabularx}
}
\end{threeparttable}
\end{table}

\begin{figure}[t]
    \centering
    \includegraphics[width=\linewidth]{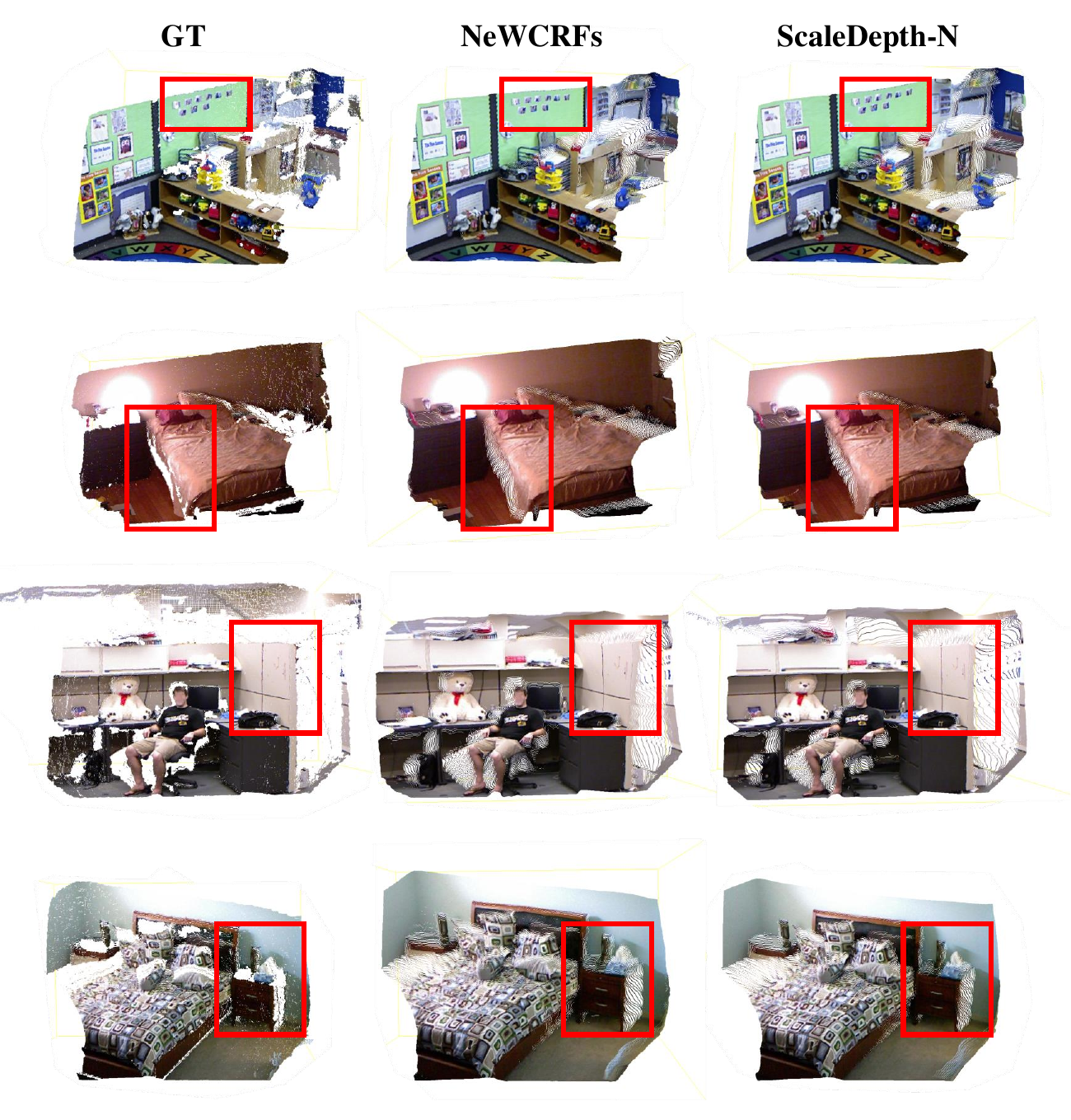}
    \caption{\textbf{The qualitative comparison of 3D point clouds reconstructed by the predicted metric depth on NYU-Depth V2 dataset. } Each row corresponds to a test sample. We use the same camera parameters to project the metric depth, and use the same viewpoints to visualize point clouds. 
    The red regions highlight that our method recovers more detailed and complete structure of the scenes.
    }
   \label{fig:pc}
\end{figure}

\textbf{Results in Indoor Scenes.}
We present the indoor evaluation results on NYU-Depth V2 dataset in~\Cref{table:nyu}.
The proposed ScaleDepth-N model, with a smaller parameter count than diffusion-based methods such as VPD~\cite{zhao2023unleashing}, significantly outperforms other state-of-the-art methods, highlighting the superiority of our architectural design.
In~\Cref{fig:nyu}, we visualize the depth maps and the error maps of our model ScaleDepth-N to provide qualitative comparisons with other methods.
The qualitative results emphasize that our method outperforms other state-of-the-art methods significantly in both global and local structural recovery. 
In particular, our method prevents overall depth shifts caused by changes in scene scale (rows 1-3) and avoids erroneously estimating the local depth distribution of the scene due to changes in lighting (rows 4-6).
To further illustrate the excellent performance of our proposed method in indoor scenes, we project the predicted depth maps into 3D space using camera parameters. 
Simultaneously, we qualitatively compare the quality of 3D point clouds reconstructed using our method with those generated by state-of-the-art methods.
As shown in~\Cref{fig:pc}, our model accurately estimates the metric depth maps of the scene, thus recovering detailed 3D structures and precise scene shapes.
Compared to NeWCRFs~\cite{yuan2022new}, our projected point clouds have fewer gaps and more complete details, highlighting the outstanding performance of our method in indoor scenes.

\textbf{Results in Outdoor Scenes.}
\Cref{table:kitti} reports the quantitative results on the outdoor benchmark KITTI.
We also visualize the qualitative results of our model ScaleDepth-K and compare them with the state-of-the-art methods as shown in~\Cref{fig:kitti_compare}.
Constrained by the sensor detection range, the depth annotations of the KITTI dataset are incomplete and biased, \ie limited to 80m. 
This limitation hinders the effective supervision of scale prediction.
Additionally, we could only use the fixed description ``outdoor scene" for all images due to the absence of category information for outdoor scenes. 
Despite lacking category labels and depth ranges, our model significantly outperforms state-of-the-art methods, highlighting its adaptability under weak conditions.

\begin{table}[t]
    \centering
    \setlength{\tabcolsep}{4pt} 
    \footnotesize
    \caption{\textbf{Quantitative results in unconstrained scenes.} Results are reported using the ARel metric. The mRI column denotes the mean relative improvement with respect to ZoeD-X-NK. Note that ZoeD-M12-NK$^\dagger$ is pretrained on extra depth datasets.}
    \label{table:unconstrained}
    \resizebox{\linewidth}{!}{
    \begin{tabularx}{1.1\linewidth}{@{}l|c|cccc|c@{}}
    \toprule
    \textbf{Method}   & \textbf{Params}        & \textbf{NYU}              & \textbf{KITTI}      & \textbf{iBims-1}                     & \textbf{vKITTI2} & \textbf{mRI} \\ 
    \midrule
    {DORN-X-NK~\cite{fu2018deep}}     &  110M  &    0.156             &    0.115       &           0.287                &    0.259           &   +53.5\%          \\
    {LocalBins-X-NK~\cite{bhat2022localbins}}  & 74M  &   0.245   &   0.133       &        0.296                   &         0.265               &  +85.0\%  \\
    {NeWCRFs-X-NK~\cite{yuan2022new}}   & 270M  &  0.109   &    0.076       &     {0.189}                      &    0.190     & +5.4\%                 \\
    {ZoeD-X-NK~\cite{bhat2023zoedepth}}    &  345M  &   {0.095}      &    {0.074}      &          \underline{0.187}                 & {0.184} & 0.0\% \\
    {ZoeD-M12-NK$^\dagger$~\cite{bhat2023zoedepth}} & 345M & \textbf{0.081} & \underline{0.061} & 0.210 & \textbf{0.112} & -14.8\% \\
    \textbf{ScaleDepth-NK}    & 216M  &   \underline{0.084}      &    \textbf{0.049}      &         \textbf{0.164}     & \underline{0.120} & -23.1\%\\  
    \bottomrule
    \end{tabularx}
    }
\end{table}

\begin{table}[t]
\centering
\caption{\textbf{Quantitative comparison on NYU and KITTI with existing methods in different experimental settings.} 
``Data'' only counts the number of labeled samples used for training.
}
\label{table:unconstrained_extra}
\resizebox{\linewidth}{!}{
    \begin{tabularx}{1.1\linewidth}{@{}l|c|*{3}{C}|*{3}{C}@{}}
        \toprule
        \multirow{2}{*}{\textbf{Method}} & \multirow{2}{*}{\textbf{Data}} 
        & \multicolumn{3}{c|}{\textbf{Outdoor (KITTI)}} & \multicolumn{3}{c}{\textbf{Indoor (NYU)}}  \\ 
                        &                & ARel$\downarrow$ & RMSE$\downarrow$ & $\delta_1\uparrow$  & ARel$\downarrow$ & RMSE$\downarrow$ & $\delta_1\uparrow$ \\ 
        \midrule
        \multicolumn{7}{l}{\textbf{Relative depth estimation methods with different training sets}} \\
        \midrule
        Midas~\cite{Ranftl2022}    & 2M   & {0.126}     & {4.609}  & {0.844} & {0.093}    & {0.386} & {0.916} \\
        DPT~\cite{ranftl2021vision}                   & 1.4M          & {0.103}     & {4.297}  & {0.894} & {0.095}    & {0.409} & {0.911} \\
        Marigold~\cite{ke2024repurposing} & 74K & 0.102 & \textbf{3.293} & 0.907 & 0.056 & 0.226 & 0.963\\
        Depth Anything~\cite{yang2024depth} & 1.5M & \textbf{0.078} &  3.320 & \textbf{0.947} & \textbf{0.042} & \textbf{0.219} & \textbf{0.982} \\
        \midrule
        \multicolumn{7}{l}{\textbf{Metric depth estimation methods with different training sets}} \\
        \midrule
        ZeroDepth~\cite{guizilini2023towards}               & 30M    & {0.102}     & {4.378}  & {0.892} & {0.100}    & {0.380} & {0.901} \\
        Metric3D~\cite{yin2023metric3d}     & 9M & \textbf{0.058} & \textbf{2.770} & \textbf{0.964} & {\textbf{0.083}} & {\textbf{0.310}} & {\textbf{0.944}} \\
        \midrule
        \multicolumn{7}{l}{\textbf{Ours: Metric depth estimation method trained on NYU and KITTI}} \\
        \midrule
        \textbf{ScaleDepth-NK}         & 47K   & \textbf{0.049}     & \textbf{2.007}  & \textbf{0.978} & \textbf{0.084}    & \textbf{0.292} & \textbf{0.940} \\ 
        \bottomrule
    \end{tabularx}
}
\end{table}

\begin{figure*}[htbp]
    \centering
    \includegraphics[width=\linewidth]{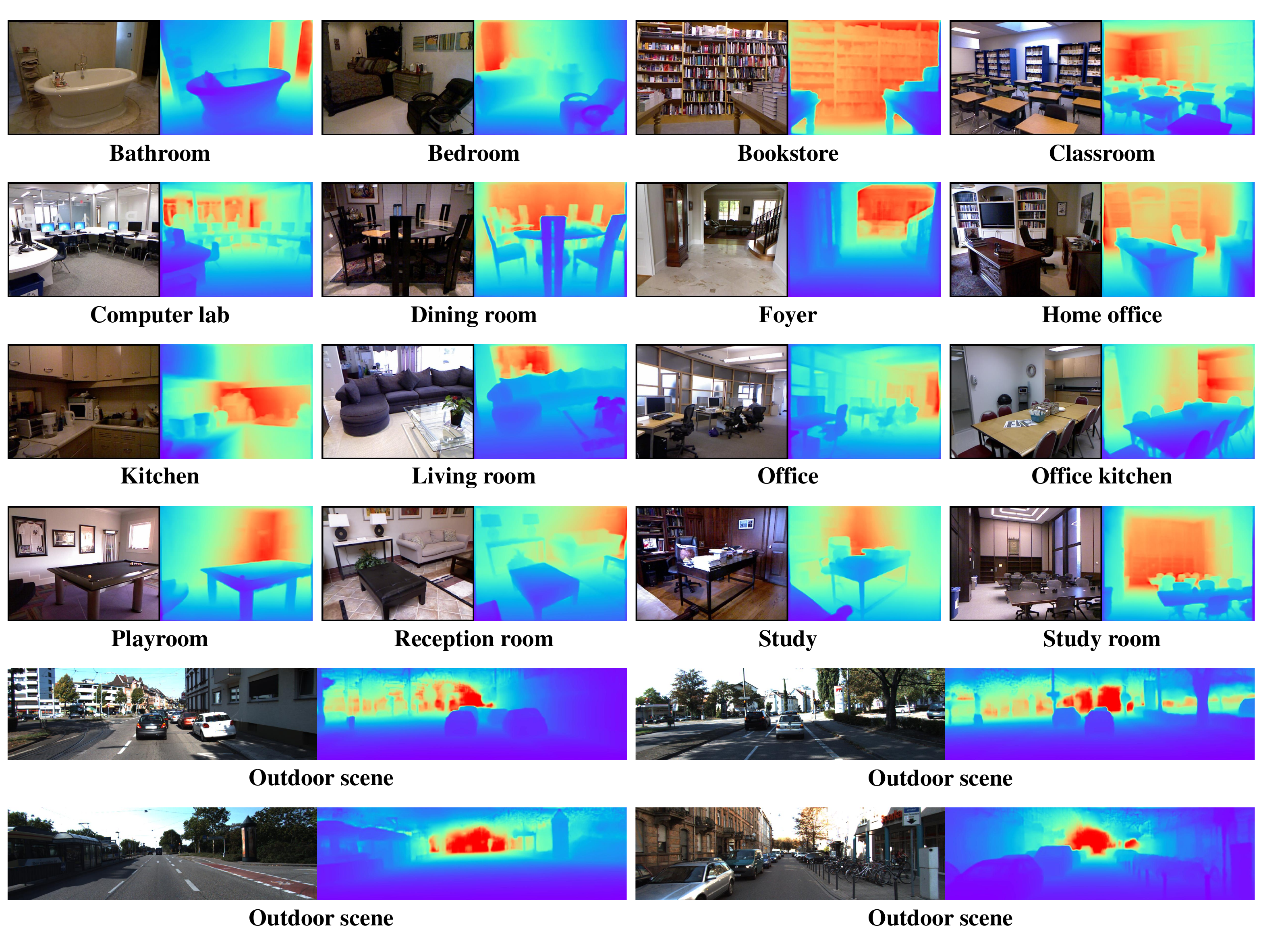}
    \caption{\textbf{The qualitative results of our model ScaleDepth-NK in unconstrained scenes. } Each couple corresponds to a test scene. Since the scene scales vary greatly, we use relative depth mapping for intuitiveness, where blue corresponds to lower \textbf{relative depth values} and red to higher values. In different categories of scenes, our method can adaptively estimate relative depth according to the depth distribution of the scene.
    }
    \label{fig:nyu_kitti}
    \vspace{3em}
    \centering
    \includegraphics[width=\linewidth]{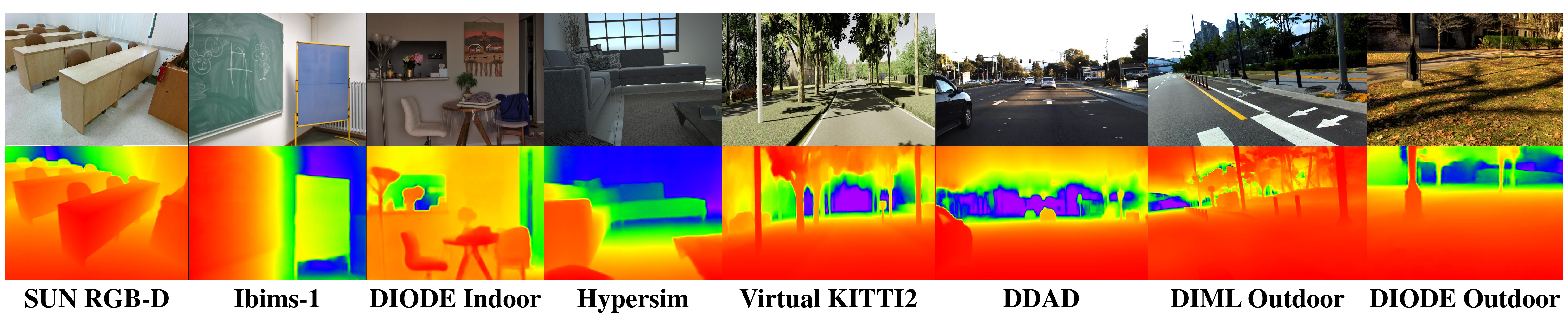}
    \caption{\textbf{The qualitative results of ScaleDepth-NK on eight unseen datasets.} 
    Since the scene scales vary greatly, we use relative depth mapping for intuitiveness, where darker colors represent greater \textbf{relative depth values}.
    Without any finetuning, our model can generalize to scenes with different scales and accurately estimate depth from indoors to outdoors.
    }
    \label{fig:zeroshot}
\end{figure*}

\textbf{Results in Unconstrained Scenes.}
The term ``unconstrained scenes" here refers to scenes of arbitrary scales. 
The considerable scale variation in unconstrained scenes, ranging from indoor to outdoor environments, presents a substantial challenge to current MDE models.
Following the experimental setup of Zoedepth~\cite{bhat2023zoedepth}, we train our model on both indoor (NYU-Depth V2) and outdoor (KITTI) datasets simultaneously, aiming to explore a universal MDE framework.
For comprehensive comparison, we compare our method with the state-of-the-art methods on both in-domain (NYU and KITTI) and out-of-domain (iBims-1 and vKITTI2) datasets with the same experimental settings.
Importantly, we employ the same parameters and configurations to evaluate the model, demanding that the model is not specifically tailored to a particular dataset or a specific scene.
As shown in~\Cref{table:unconstrained}, our method outperforms the state-of-the-art method ZoeDepth-X-NK by 23.1\% (mean relative improvement in ARel), highlighting the superiority of the proposed approach. 
The visualization results are shown in~\Cref{fig:nyu_kitti}, where our methods predict accurate depth in both indoor and outdoor scenes.
Meanwhile, we also note that some recent methods have achieved impressive results by training models with amazing amounts of data. 
Although the experimental settings are different, we provide comparisons with them in~\Cref{table:unconstrained_extra} for reference, where our method even surpasses some methods trained in large-scale depth datasets.

\begin{figure*}[t]
    \begin{minipage}[t]{\linewidth}
        \centering
        \includegraphics[width=\linewidth]{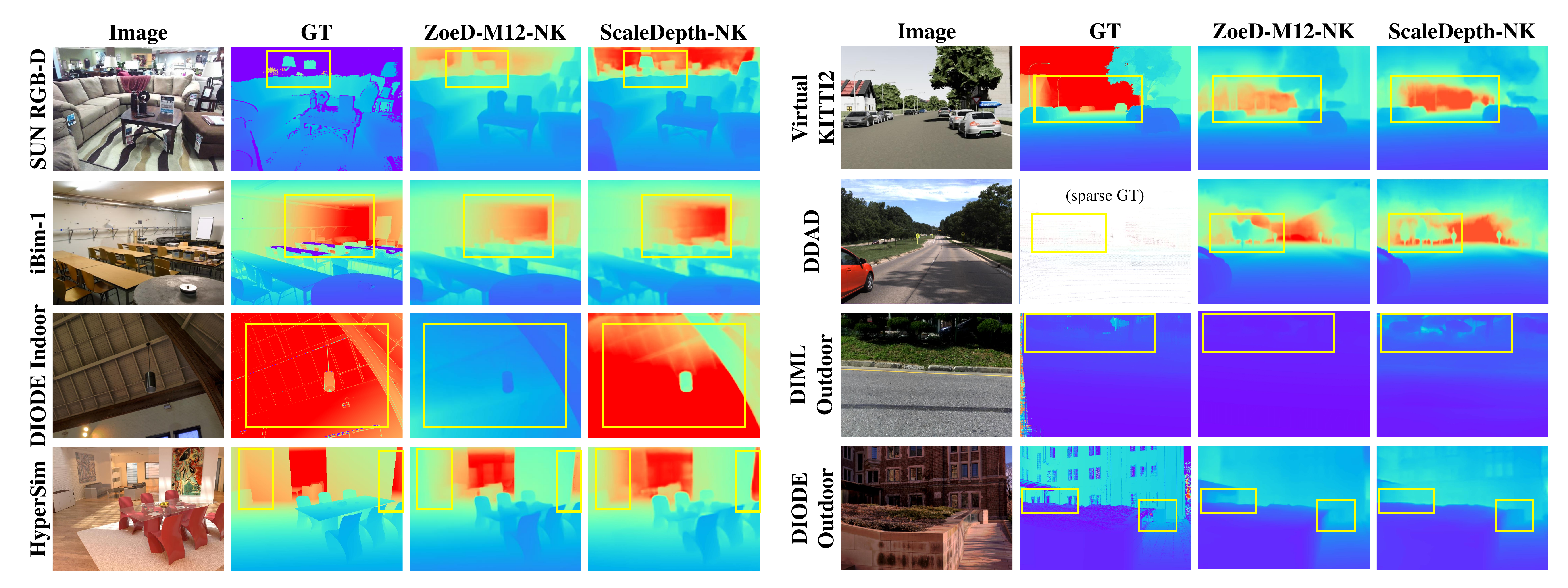}
        \caption{\textbf{The zero-shot evaluations in unseen scenes. } 
        Each row corresponds to a test sample. For indoor scenes, we set the depth range of the colormap to 0-10m; for outdoor scenes, we set the depth range of the colormap to 0-80m. Blue corresponds to lower \textbf{metric depth values} and red to higher values. Purple represents the invalid area in ground truth (GT). The yellow box highlights that our method is capable of predicting more accurate metric depth values.
        }
        \label{fig:zeroshot_all}        
    \end{minipage}
\end{figure*}

\textbf{Results in Unseen Scenes.}
To demonstrate the generalization capability of our model, we report the zero-shot evaluation results on eight unseen datasets as shown in~\Cref{table:unseen_indoor,table:unseen_outdoor}.
To emphasize the zero-shot generalization performance of our model, we also provide a qualitative result as shown in~\Cref{fig:zeroshot} and comparison with the state-of-the-art method ZoeD-M12-NK~\cite{bhat2023zoedepth} as shown in~\Cref{fig:zeroshot_all}.
Notably, ScaleDepth-N/K permits fair comparison with ZoeD-X-N/K, where our models significantly outperform them on eight unseen datasets, demonstrating its strong zero-shot generalization ability.
However, ZoeD-M12-NK has obvious advantages in the zero-shot test due to the use of extra depth datasets for pre-training.
For example, the DIML Outdoor dataset contains many out-of-distribution scenes, such as construction sites and parks. 
These scene categories might appear in the pre-training dataset of ZoeD-M12-NK, which our model has never seen before.
Nevertheless, our model ScaleDepth-NK still exhibits competitive or even better performance on other seven unseen datasets compared with ZoeD-M12-NK, demonstrating its superior zero-shot performance.
Moreover, different from ZoeDepth~\cite{bhat2023zoedepth}, our method do not require setting depth range manually during training or testing.

\begin{table*}[t]
\centering
\caption{\textbf{Quantitative results in unseen indoor scenes.} The method with $\dagger$ is trained on extra depth datasets. 
The maximum depth of evaluation is set to 8m for SUN RGB-D, 10m for iBims-1 and DIODE Indoor, and 80m for HyperSim.
}
\label{table:unseen_indoor}
\resizebox{\linewidth}{!}{
\begin{tabularx}
{\linewidth}
{@{}
    l|
    *{3}{C}|
    *{3}{C}|
    *{3}{C}|
    *{3}{C}
    @{}}
\toprule
             \multirow{2}{*}{Method} & \multicolumn{3}{c|}{\textbf{SUN RGB-D}}                         & \multicolumn{3}{c|}{\textbf{iBims-1 Benchmark}}                 & \multicolumn{3}{c|}{\textbf{DIODE Indoor}}                      & \multicolumn{3}{c}{\textbf{HyperSim}}                           \\
& $\delta_1\uparrow$ & ARel$\downarrow$ & RMSE$\downarrow$ & $\delta_1\uparrow$ & ARel$\downarrow$ & RMSE$\downarrow$ & $\delta_1\uparrow$ & ARel$\downarrow$ & RMSE$\downarrow$ & $\delta_1\uparrow$ & ARel$\downarrow$ & RMSE$\downarrow$ \\ \midrule
BTS~\cite{lee2019big} & 0.740                   & 0.172              & 0.515            & 0.538                   & 0.231              & 0.919            & 0.210                   & 0.418              & 1.905            & 0.225                   & 0.476              & 6.404            \\
AdaBins~\cite{bhat2021adabins}      & 0.771                   & 0.159              & 0.476            & 0.555                   & 0.212              & 0.901            & 0.174                   & 0.443              & 1.963            & 0.221                   & 0.483              & 6.546            \\
NeWCRFs~\cite{yuan2022new} & 0.798                   & 0.151              & 0.424            & 0.548                   & 0.206              & 0.861            & 0.187                   & 0.404              & 1.867            & 0.255                   & 0.442              & 6.017            \\
ZoeD-X-N~\cite{bhat2023zoedepth} & \underline{0.857}                   & \textbf{0.124}              & \underline{0.363}            & \underline{0.668}                   & \underline{0.173}              & \underline{0.730}            & \underline{0.400}                   & \underline{0.324}             & \underline{1.581}            & \underline{0.284}                   & \underline{0.421}              & \underline{5.889}            \\
\textbf{ScaleDepth-N} & \textbf{0.864} & \underline{0.127} &  \textbf{0.360}          & \textbf{0.788}  & \textbf{0.156}  & \textbf{0.601} &  \textbf{0.455}  &  \textbf{0.277}  &  \textbf{1.350} & \textbf{0.383} &  \textbf{0.393} &  \textbf{5.014}  \\ 
\midrule
ZoeD-M12-NK$\dagger$~\cite{bhat2023zoedepth} & {0.856} &  {\textbf{0.123}} & {\textbf{0.356}}  & 
{0.615} & {0.186} & {0.777}  & {0.386} &  {\textbf{0.331}} &   {1.598}  &   {0.274} &    {0.419} & {5.830}  \\
\textbf{ScaleDepth-NK} &  \textbf{0.866}  &   0.129   &  {0.359}      &  \textbf{0.778}   &  \textbf{0.164}  &  \textbf{0.590}   &   \textbf{0.447}  &   0.355   &   \textbf{1.443}  &  \textbf{0.413}    &  \textbf{0.381}   &  \textbf{4.825}   \\ 
\bottomrule
\end{tabularx}
}
\vspace{1em}

\centering
\caption{\textbf{Quantitative results in unseen outdoor scenes.} The method with $\dagger$ is trained on extra depth datasets. 
The maximum depth of evaluation is set to 80m for all unseen outdoor datasets.
}
\label{table:unseen_outdoor}
\resizebox{\linewidth}{!}{
\begin{tabularx}
{\linewidth}
{@{}
    l|
    *{3}{C}|
    *{3}{C}|
    *{3}{C}|
    *{3}{C}
    @{}}
\toprule
             \multirow{2}{*}{Method} & \multicolumn{3}{c|}{\textbf{Virtual KITTI 2}}                         & \multicolumn{3}{c|}{\textbf{DDAD}}                 & \multicolumn{3}{c|}{\textbf{DIML Outdoor}}                      & \multicolumn{3}{c}{\textbf{DIODE Outdoor}}                           \\
& $\delta_1\uparrow$ & ARel$\downarrow$ & RMSE$\downarrow$ & $\delta_1\uparrow$ & ARel$\downarrow$ & RMSE$\downarrow$ & $\delta_1\uparrow$ & ARel$\downarrow$ & RMSE$\downarrow$ & $\delta_1\uparrow$ & ARel$\downarrow$ & RMSE$\downarrow$ \\ \midrule
BTS~\cite{lee2019big} & 0.831 & 0.115 & 5.368  & 0.805 & 0.147 & 7.550 & \underline{0.016} & 1.785 & \underline{5.908} & 0.171 & 0.837 & 10.48 \\
AdaBins~\cite{bhat2021adabins} & 0.826 & 0.122 & 5.420 & 0.766 & 0.154 & 8.560 & 0.013 & 1.941 & 6.272 & 0.161 & 0.863 & 10.35 \\
NeWCRFs~\cite{yuan2022new} &  0.829 & 0.117 & 5.691 & \textbf{0.874} & \textbf{0.119} & \textbf{6.183} & 0.010 & 1.918 & 6.283  & 0.176 & 0.854 & 9.228  \\
ZoeD-X-K~\cite{bhat2023zoedepth} & \underline{0.837} & \underline{0.112} & \underline{5.338} & 0.790 & 0.137 & 7.734 & 0.005 & \underline{1.756} & 6.180 & \underline{0.242} &  \underline{0.799} & \underline{7.806}  \\
\textbf{ScaleDepth-K} &  \textbf{0.882}  & \textbf{0.096}  & \textbf{4.495}  &   \underline{0.863} & \underline{0.120}  &  \underline{6.378} & \textbf{0.033} &  \textbf{1.437} &  \textbf{4.910}    &  \textbf{0.333}  & \textbf{0.605}  & \textbf{6.950}  \\
\midrule
ZoeD-M12-NK$\dagger$~\cite{bhat2023zoedepth}  &  {\textbf{0.850}} & {\textbf{0.105}} &   {5.095}  &  {0.824} & {0.138} & {7.225}  & {\textbf{0.292}} & {\textbf{0.641}} & {\textbf{3.610}} & {0.208} & {0.757} & {\textbf{7.569}} \\
\textbf{ScaleDepth-NK} &  0.834  & 0.120  & \textbf{4.747}    &   \textbf{0.871}               & \textbf{0.121}     &  \textbf{6.097}    &  {0.058}     &   {1.007}    &   {4.344}   &  \textbf{0.262}    &   \textbf{0.562}   &  {8.632}      \\
 \bottomrule
\end{tabularx}
}
\end{table*}

\subsection{Ablation Study}
\label{sec:ablationstudy}

\begin{table*}[t]
\centering
\caption{\textbf{Ablation study about different model components in unconstrained scenes.} 
ARel and RMSE reflect the performance of the metric depth, while SILog reflects the performance of relative depth.
We set the depth range to 0-80m for models [A], [B], [C], while model [D] and [E] do not require setting depth range during training.
}
\label{table:ablation_nyukitti}
\resizebox{\linewidth}{!}{
    \begin{tabularx}{\linewidth}{@{\hspace{3pt}}c@{\hspace{3pt}}|@{\hspace{6pt}}c@{\hspace{6pt}}c@{\hspace{6pt}}c@{\hspace{6pt}}|*{3}{C}|*{3}{C}@{}}
        \toprule
        \multirow{2}{*}{Model} &   \multirow{2}{*}{Depth Discretization} & \multirow{2}{*}{Mask Generation} & \multirow{2}{*}{Scale Prediction}
        & \multicolumn{3}{c|}{\textbf{Outdoor (KITTI)}} & \multicolumn{3}{c}{\textbf{Indoor (NYU)}}  \\ 
                        &           &        &        & ARel$\downarrow$ & RMSE$\downarrow$ & SILog$\downarrow$  & ARel$\downarrow$ & RMSE$\downarrow$ & SILog$\downarrow$ \\ \midrule
        A       &  &  &  & 0.056   & 2.186  & 7.655 & 0.116    & 0.351 & 9.485 \\
        B       & $\checkmark$   &   &      &  0.054 & 2.092  &  7.294 & 0.087   & 0.319 & 8.571 \\
        C       & $\checkmark$  & $\checkmark$  &         & 0.051     & {2.103}  & {7.091} & 0.086    & 0.299 & 8.444 \\
        D       & $\checkmark$  &   &  $\checkmark$       & 0.053     & {2.019}  & {6.906} & 0.084    & 0.294 & 8.221 \\        
        E       & $\checkmark$  &  $\checkmark$ & $\checkmark$  & \textbf{0.049}     & \textbf{2.007}  & \textbf{6.776} & \textbf{0.084}    & \textbf{0.292} & \textbf{8.145} \\ \bottomrule
    \end{tabularx}
}
\end{table*}

In this section, we study the impact of different designs on model performance.

\textbf{Model Components.}
To investigate the effectiveness of the model components, we adopt five different configurations of models and compare their performance in unconstrained scenes. 
We add a depth regression head directly after the image encoder as our baseline, denoted as model [A].
By dividing the depth bins, we design the bin query to adaptively predict the depth distribution, denoted as model [B]. 
By adding mask generation, we can use mask attention to let the bin query interact with the corresponding image features, denoted as model [C]. 
Finally, we add scale query to adaptively predict the scene scale based on model [B] and model [C], denoted as model [D] and model [E].
The experimental results are shown in~\Cref{table:ablation_nyukitti}, revealing the following insights:
(1) Compared to the continuous regression-based model [A], the discrete regression-based model [B] achieves significant performance improvements by adopting depth discretization (split depth range to bins). 
(2) The mask generation mechanism in the ADRE module encourages model [C] to focus on depth-related regions and improve feature aggregation, further enhancing the model performance.
(3) 
By explicitly modeling the scale of the scene, we no longer need to set fixed depth ranges during training. 
Instead, we let the model adaptively estimate the scale of each scene. 
With the assistance of the SASP module, model [D] outperforms model [B] in both the metric depth and the relative depth estimation. 
(4)
Through metric depth decomposition, model [E] effectively integrates the scale prediction and relative depth estimation branches into a unified framework, thus achieving the state-of-the-art performance. 
%
%

\textbf{Text Prompts for Scale Prediction.}
To demonstrate the effectiveness of incorporating semantic information into the SASP module, we conduct ablation experiments in unconstrained scenes with different conditions. 
As shown in~\Cref{table:text_prompts}, we compare the ARel metric of models with no scale prediction, image conditioned, and text conditioned.
We find that using text prompts as condition can lead to greater performance gains.
Indeed, the benefits of semantic supervision are evident. 
The proposed semantic supervision aligns scale queries with text features, implicitly injecting semantic information of the scene to achieve more precise scale prediction with no extra cost during inference. 
When text prompts are not available, the SASP module degrades to inferring scales based on the features of each image during training (image conditioned).
Despite this, our model still outperforms the current state-of-the-art methods in unconstrained scenes under same experimental settings.

\begin{figure*}[t]
  \centering
   \includegraphics[width=\linewidth]{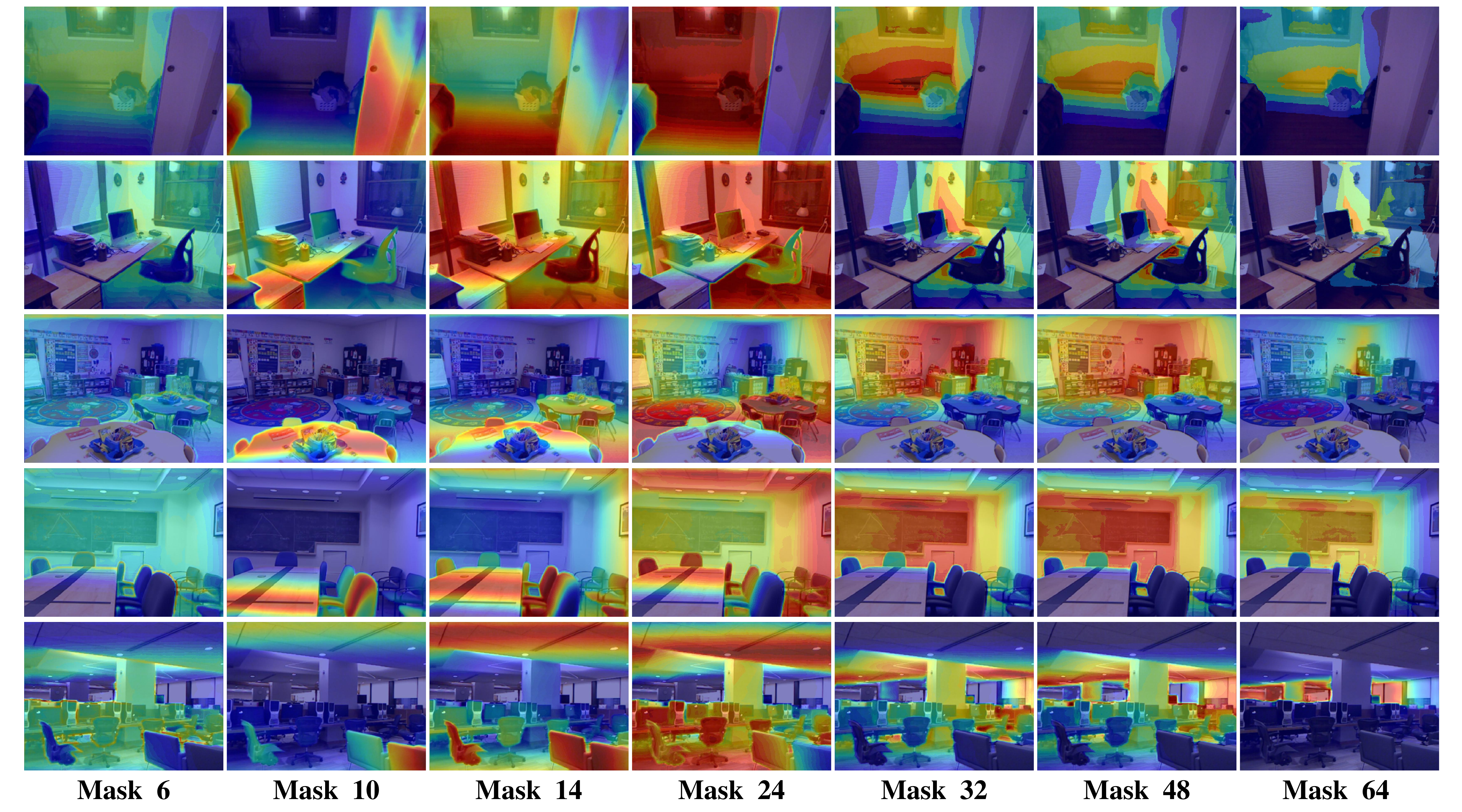}
    \vspace{-2em}
   \caption{\textbf{The attention masks of bin queries.} The $i$-th mask represents the similarity between $i$-th bin features and image features, which is also the attention mask of the transformer in the next layer. The red region denotes higher similarity, while the blue region indicates lower similarity. The visualization results demonstrate that the designed bin queries can adaptively estimate relative depth based on the depth distribution of the scene.
   }
   \label{fig:qam}
\end{figure*}

\begin{table}[t]
\centering
\caption{\textbf{Ablation study about different conditions in unconstrained scenes.}  We ablate our model ScaleDepth-NK on both KITTI and NYU-Depth V2 dataset,}
\label{table:text_prompts}
\begin{threeparttable}
\resizebox{\linewidth}{!}{
\begin{tabular}{c|ccc|ccc}
    \toprule
    \multirow{2}{*}{Setting} & \multicolumn{3}{c|}{\textbf{Outdoor (KITTI)}}  & \multicolumn{3}{c}{\textbf{Indoor (NYU)}} \\
      & ARel$\downarrow$ & RMSE$\downarrow$ & SIlog$\downarrow$ & ARel$\downarrow$ & RMSE$\downarrow$ & SILog$\downarrow$ \\ \midrule
    no scale prediction & 0.051 & 2.103 & 7.091 & 0.086 & 0.299 & 8.444 \\
    image conditioned   &  0.051 & \textbf{2.004} & 6.830 & 0.090 & 0.293 &  8.162 \\
    text conditioned  & \textbf{0.049} & {2.007} & \textbf{6.776} & \textbf{0.084}  & \textbf{0.292} & \textbf{8.145}\\
    \bottomrule
\end{tabular}
}
\end{threeparttable}
\end{table}
\begin{table}[t]
\centering
\caption{\textbf{Ablation study about various backbones.} We ablate our model ScaleDepth-N on NYU-Depth V2 dataset.}
\label{table:backbone}
\begin{threeparttable}
\resizebox{0.8\linewidth}{!}{
\begin{tabular}{c|cccc}
    \toprule
    Backbone             & ARel$\downarrow$ & RMSE$\downarrow$  & SILog$\downarrow$ & $\delta_1\uparrow$ \\ \midrule
    Swin-L             & 0.088  & 0.318 & 9.009  & 0.934 \\
    ConvNext-L             & 0.086  & 0.310 & 8.822 & 0.939 \\
    Frozen CLIP          & 0.084    & 0.298 & 8.778 & 0.941 \\    
    CLIP        & \textbf{0.074}    & \textbf{0.267} & \textbf{7.626} & \textbf{0.957} \\
    \bottomrule
\end{tabular}
}
\end{threeparttable}
\end{table}

\begin{table*}[htbp]
\centering
\caption{\textbf{Ablation study about models with different backbones in unseen scenes.}  We train our model ScaleDepth-N on NYU-Depth V2 dataset and zero-shot test it on four unseen indoor dataset.
}
\label{table:ablation_unseen_indoor}
\resizebox{\linewidth}{!}{
\begin{tabularx}
{\linewidth}
{@{}
    l|c|
    *{3}{C}|
    *{3}{C}|
    *{3}{C}|
    *{3}{C}
    @{}}
\toprule
             \multirow{2}{*}{Method} & \multirow{2}{*}{Backbone}& \multicolumn{3}{c|}{\textbf{SUN RGB-D}}                         & \multicolumn{3}{c|}{\textbf{iBims-1 Benchmark}}                 & \multicolumn{3}{c|}{\textbf{DIODE Indoor}}                      & \multicolumn{3}{c}{\textbf{HyperSim}}                           \\
& & $\delta_1\uparrow$ & ARel$\downarrow$ & RMSE$\downarrow$ & $\delta_1\uparrow$ & ARel$\downarrow$ & RMSE$\downarrow$ & $\delta_1\uparrow$ & ARel$\downarrow$ & RMSE$\downarrow$ & $\delta_1\uparrow$ & ARel$\downarrow$ & RMSE$\downarrow$ \\ \midrule
NeWCRFs~\cite{yuan2022new} & Swin-L & 0.798                   & 0.151              & 0.424            & 0.548                   & 0.206              & 0.861            & 0.187                   & 0.404              & 1.867            & 0.255                   & 0.442              & 6.017            \\
ZoeD-X-N~\cite{bhat2023zoedepth} & BEiT-L & \underline{0.857}                   & \textbf{0.124}              & \underline{0.363}            & 0.668                   & {0.173}              & {0.730}            & {0.400}                   & {0.324}             & {1.581}            & {0.284}                   & {0.421}              & {5.889}            \\
\textbf{ScaleDepth-N} & Swin-L & 0.822 & 0.144 & 0.410  & \underline{0.750}  &  \underline{0.168} & \underline{0.670} & \underline{0.425} &  \underline{0.306}  & \underline{1.476}  & \underline{0.342} & \underline{0.420} & \underline{5.284} \\
\textbf{ScaleDepth-N} & CLIP & \textbf{0.864} & \underline{0.127} &  \textbf{0.360}          & \textbf{0.788}  & \textbf{0.156}  & \textbf{0.601} &  \textbf{0.455}  &  \textbf{0.277}  &  \textbf{1.350} & \textbf{0.383} &  \textbf{0.393} &  \textbf{5.014}  \\ 
\bottomrule
\end{tabularx}
}
\end{table*}

\textbf{Mask Generation for Relative Depth Estimation.}
To further demonstrate the effectiveness of the mask generation mechanism, we visualize the attention masks corresponds to different bins in~\Cref{fig:qam}.
Since bin queries estimate relative depth with a normalized 0-1 depth space, they can adaptively adjust the activated regions based on the depth distributions of different images.
Through masked attention, each bin query focus on its own depth-related regions, thereby achieving better local structure recovery and relative depth estimation. 

\textbf{Pretrained Image Encoders.}
We observe that the pre-trained image encoder plays a crucial role in the performance of the model.
In this work, we investigate the impact of model pretraining in various ways on depth estimation in~\Cref{table:backbone}.
%
%
Although the compared backbones have similar parameter size, the experiment indicates that the fine-tuned CLIP model yields the greatest performance, significantly exceeding Swin-Large~\cite{liu2021Swin}, ConvNext-Large~\cite{liu2022convnet} or frozen CLIP.
We argue that although the pre-trained CLIP model provides a good model initialization, it still requires fine-tuning for depth estimation task due to the inherent differences between semantic and depth features.
Meanwhile, We notice that using CLIP as our backbone may have advantages in comparison with other methods.
Therefore, we replace the CLIP image encoder with Swin-L and retrain our model ScaleDepth-N. 
The zero-shot generalization results are shown in~\cref{table:ablation_unseen_indoor}, where the retrained model still exhibits state-of-the-art performance on unseen datasets, highlighting the effectiveness of the proposed architecture.

\begin{figure}[t]
  \centering
   \includegraphics[width=0.8\linewidth]{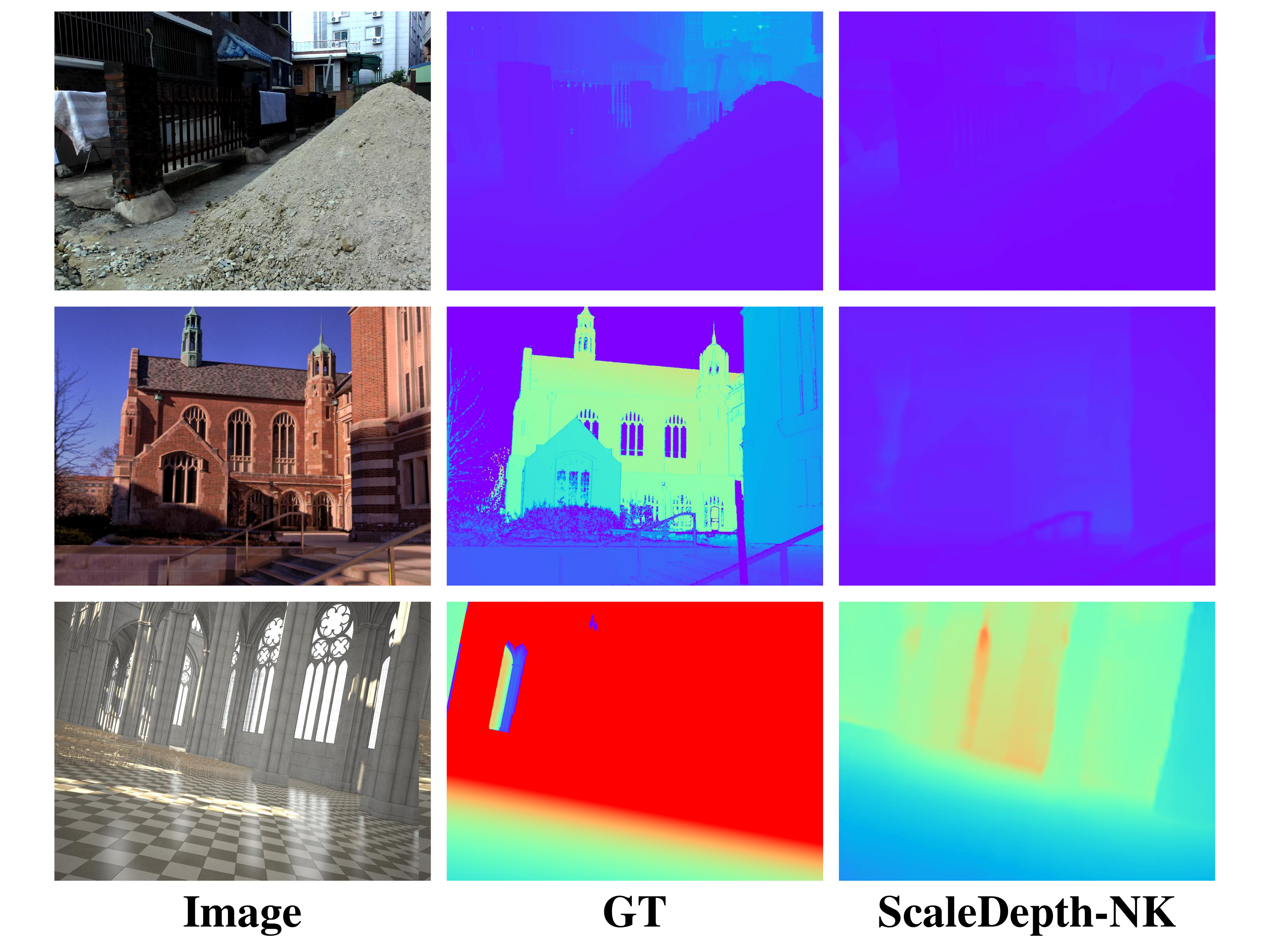}
   \caption{\textbf{The failure cases in unseen scenes. } Each row corresponds to one test sample for zero-shot evaluations of our model ScaleDepth-NK. The depth maps are mapped by using color map \textit{rainbow}. Blue corresponds to lower \textbf{metric depth values} and red to higher values.
   Since our model has never seen mounds, outdoor buildings, or palaces at all during training, it exhibits a certain scale shift in the zeroshot test.
}
   \label{fig:fail}
\end{figure}

\section{Limitations}
\label{sec:limit}
While the CLIP model~\cite{pmlr-openclip} exhibits strong zero-shot generalization capabilities across many downstream tasks without fine-tuning, so far it cannot be directly applied to depth estimation tasks. 
Although our method achieves accurate metric depth estimation in indoor and outdoor scenes through fine-tuning the CLIP model, there may still be failures in some unseen scenes. 
As shown in~\Cref{fig:fail}, if the tested scenes belong to completely unseen categories, our model may struggle to predict precise scene scale. 
Therefore, we believe that training the model on a more extensive range of scenes could further enhance its generalization ability.
The proposed method explores how to achieve accurate metric depth estimation in scenes of different scales. 
In future research, it would be valuable to investigate how to develop a universal depth estimation model that can zero-shot generalize to diverse scenes in real-world applications.

\section{Conclusion}
\label{sec:Conclusion}

In this work, we propose a novel metric depth estimation method ScaleDepth, which decomposes metric depth estimation into scale prediction via an SASP module and relative depth estimation via an ARDE module.
%
With these two well-designed modules, our ScaleDepth achieves both indoor and outdoor depth estimation in a unified framework. 
Extensive results under four kinds of experimental settings demonstrate the superiority of the proposed method.
In future work, we would like to explore a universal MDE framework, which can predict accurate metric depth in an open-vocubulary world.

\section*{Acknowledgments}
We thank Jianfeng He and Jiacheng Deng for their thoughtful and valuable suggestions.

{\appendix[Model Details]
\label{sec:model}
In this section, we provide more details about the components of our models. 

\textbf{Image Encoder.}
We adopt the CLIP~\cite{pmlr-openclip} backbone pretrained on LAION-2B~\cite{schuhmann2022laion} as our image encoder.
We extract multilevel visual tokens from CLIP instead of class tokens as image features, as they retain more specific structural information of the image.
As shown in our ablation study, we find that fine-tuning CLIP is the simplest and most effective ways to incorporate CLIP into depth estimation.
To retain pre-trained knowledge, we set the learning rate of the whole image encoder as 1/10 of the base learning rate.

\textbf{Text Encoder.}
We design text prompts to obtain the text embeddings of different scene categories by a frozen CLIP text encoder.
The text prompts are based on manually designed templates, and~\Cref{table:text} provides some examples of them.
For each scene category, we take the average of text features extracted from all templates as the text embeddings.
%
Naturally, we use the pre-trained CLIP model with aligned text and image encoders, where its aligned image encoder and text encoder can facilitate fast convergence and stable training. 

\begin{table}
\begin{minipage}[t]{\linewidth}
\centering
\caption{\textbf{Examples of text templates and scene categories.} 80 text templates and 28 scene categories are used in our experiments to generate $80\times28$ text prompts.}
\label{table:text}
\setlength\tabcolsep{6pt}
\small
    \begin{tabular}{c}
        \toprule
        Text templates \\
        \midrule
        a low resolution photo of a [class].       \\
        a bad photo of the [class].        \\
        a cropped photo of the [class].     \\
        a bright photo of a [class].         \\
        a photo of a clean [class].        \\
        a photo of the dirty [class].      \\
        a good photo of the [class].       \\
        a photo of one [class].            \\
        ... \\
        a photo of a large [class].          \\
        \bottomrule
    \end{tabular}    
    \hspace{1em}
    \begin{tabular}{c}
        \toprule
        Scene categories \\
        \midrule
        printer room      \\
        bathroom      \\
        living room      \\
        conference room      \\
        study room      \\
        kitchen      \\
        bedroom      \\
        dinette      \\
        ... \\
        outdoor scene      \\
        \bottomrule
    \end{tabular}     
\end{minipage}
\end{table}

\begin{table}
\begin{minipage}[t]{\linewidth}
\centering
\caption{\textbf{Multi-level features generated by pixel decoder.}  The input image is randomly cropped to $480\times480$ during training on the NYU-Depth V2 dataset.}
\label{table:pixel}
\begin{threeparttable}
\setlength\tabcolsep{6pt}
\small
    \begin{tabular}{c|c|cc}
        \toprule
        Backbone & Level & Before pixel decoder & After pixel decoder \\
        \midrule
        \multirow{4}{*}{CLIP}   & 1  & $120\times120\times192$  & $120\times120\times256$ \\
                                & 2  & $60 \times60 \times384$  & $60 \times60 \times256$ \\
                                & 3  & $30 \times30 \times768$  & $30 \times30 \times256$ \\
                                & 4  & $15 \times15 \times1536$ & $15 \times15 \times256$ \\
        \bottomrule
    \end{tabular}
\end{threeparttable}
\end{minipage}
\end{table}

\textbf{Pixel Decoder.}
The pixel decoder is adopted from Mask2Former~\cite{cheng2021mask2former}, which upsamples low-resolution features from the image encoder to generate high-resolution per-pixel embeddings.
In our implementation, we use four levels of features with different resolutions.
\Cref{table:pixel} provide the detailed information of the decoded image features during training on the NYU-Depth V2 dataset.

\textbf{Transformer Layers.}
We use a 9-layer transformer in the decoder network, which can be divided into 3 blocks, with each block containing 3 layers.
In each block, we adopt a multi-scale deformable attention to gradually aggregate last 3 levels of the image features generated by the pixel decoder.
The bin queries and the scale queries are then passed into masked attention to interact with the image features at each layer.
For bin queries, we use the generated mask in the ADRE module as the attention mask.
For the scale queries, we aim to capture global semantic information for the semantic-aware scale prediction.
Therefore, we do not set masks so that the scale queries can interact fully with the image features.
The first-level features are then used to calculate the similarity of bin features and image features in the ARDE module.
%

\textbf{Computational Complexity.}
The computational complexity analysis of each component of ScaleDepth-N is shown in~\Cref{table: overhead}. 
As described in the paper, our model ScaleDepth-N significantly outperforms other models with similar parameter size on the NYU-Depth V2 dataset. 
Additionally, under same experimental settings, our model ScaleDepth-NK achieves superior results in both indoor and outdoor scenes with a smaller parameter size compared with Zoedepth-X-NK~\cite{bhat2023zoedepth}, highlighting the effectiveness of the proposed approach.

\begin{table}[htbp]
\begin{minipage}[t]{\linewidth}
\centering
\caption{\textbf{Computational complexity analysis.} 
We test our ScaleDepth-N on an Nvidia RTX 3090 with an input image of size $640\times480$. }
\label{table: overhead}
\begin{threeparttable}
\setlength\tabcolsep{6pt}
\small
    \begin{tabular}{c|c|cc}
        \toprule
        Components & Params & FLOPs \\
        \midrule
        Image encoder               &  196.0M     &  158.0G  \\
        Pixel decoder               &  5.8M    &  31.6G \\
        Transformer Layers          &  14.2M    &  3.0G \\
        Miscs                       &  0.2M     &  0.1G  \\
        \midrule
        Total                       &  216.2M     &  192.7G \\
        \bottomrule
    \end{tabular}
\end{threeparttable}
\end{minipage}
\end{table}

}

\bibliographystyle{IEEEtran}
\bibliography{main}

\end{document}